\documentclass{article}
\usepackage[dvipsnames,table]{xcolor}

\usepackage{microtype}
\usepackage{graphicx}
\usepackage{subfigure}
\usepackage{booktabs} 

\usepackage{hyperref}

\usepackage[accepted]{icml2024}

\usepackage{amsmath}
\usepackage{amssymb}
\usepackage{mathtools}
\usepackage{amsthm}
\usepackage{lipsum}
\usepackage{listings}
\usepackage{multirow}
\usepackage{longtable}

\usepackage[capitalize,noabbrev]{cleveref}

\theoremstyle{plain}

\theoremstyle{definition}

\theoremstyle{remark}

\usepackage[textsize=tiny]{todonotes}

\icmltitlerunning{\method: Execution-Free Repository-Grounded Plan Search for Code-Use}

\newcommand{\AllSymbols}{\mathcal{B}}         
\newcommand{\AllStrings}{\mathcal{L}}         
\newcommand{\AllPlans}{\mathcal{P}}           

\newcommand{\intent}{t}                       
\newcommand{\relevantSymbols}{B}             
\newcommand{\planStep}{x}                     
\newcommand{\plan}{p}                         

\newcommand{\intentToSymbols}{g}             
\newcommand{\mutationFn}{s}                
\newcommand{\rankingFn}{h}                   

\newcommand{\stepTuple}{(\intent, \relevantSymbols)}  
\newcommand{\planSequence}{[x_1, ..., x_n]}          

\newcommand{\lca}{LongCodeArena}
\newcommand{\method}[0]{\textsc{MutaGReP}}

\LTcapwidth=\textwidth 

\begin{document}

\twocolumn[
\icmltitle{\method: Execution-Free Repository-Grounded Plan Search for Code-Use}


\icmlsetsymbol{equal}{*}

\begin{icmlauthorlist}
\icmlauthor{Zaid Khan}{unc}
\icmlauthor{Ali Farhadi}{ai2}
\icmlauthor{Ranjay Krishna}{ai2}
\icmlauthor{Luca Weihs}{vercept}
\icmlauthor{Mohit Bansal}{unc}
\icmlauthor{Tanmay Gupta}{ai2}
\end{icmlauthorlist}

\icmlaffiliation{ai2}{Allen Institute for Artificial Intelligence (Ai2)}
\icmlaffiliation{unc}{University of North Carolina, Chapel Hill}
\icmlaffiliation{vercept}{Vercept AI (work done while at Ai2)}

\icmlcorrespondingauthor{Zaid Khan}{zaidkhan@cs.unc.edu}
\icmlcorrespondingauthor{Tanmay Gupta}{tanmayg@allenai.org}

\icmlkeywords{LLM-guided search,LLM-guided planning,repo-level code use,natural language plan search,test-time compute,code generation}

\vskip 0.3in
]



\printAffliationsNoNotice{}

\begin{abstract}
When a human requests an LLM to complete a coding task using functionality from a large code repository, how do we provide context from the repo to the LLM?
One approach is to add the entire repo to the LLM's context window.
However, most tasks involve only fraction of symbols from a repo, longer contexts are detrimental to the LLM's reasoning abilities~\cite{babilong}, and context windows are not unlimited.
Alternatively, we could emulate the human ability to navigate a large repo, pick out the right functionality, and form a plan to solve the task.
We propose \method{} (\textbf{Muta}tion-guided \textbf{G}rounded \textbf{Re}pository \textbf{P}lan Search), an approach to search for plans that decompose a user request into natural language steps grounded in the codebase. \method~performs neural tree search in plan space, exploring by mutating plans and using a symbol retriever for grounding. 
On the challenging \lca~benchmark, our plans use less than $5\%$ of the 128K context window for GPT-4o but rival the coding performance of GPT-4o with a context window filled with the repo.
Plans produced by \method~allow Qwen 2.5 Coder 32B and 72B to match the performance of GPT-4o with full repo context and enable progress on the hardest \lca~tasks. Project page: \href{https://zaidkhan.me/MutaGReP}{zaidkhan.me/MutaGReP}
\end{abstract}

\section{Introduction}
\label{sec:introduction}
\begin{figure}[!ht]
    \centering
    \includegraphics[width=1\linewidth]{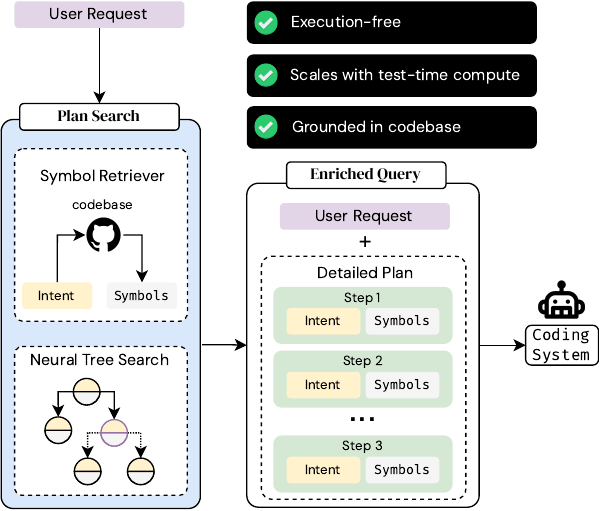}
    \vspace{-3mm}
    \caption{\small\textbf{\method~Overview} 
    Given a user request that requires writing code against a specific codebase, we search for realizable plans to solve the user's request using LLM-guided tree search. Our search procedure uses a symbol retriever to constrain search to plans which are implementable with symbols available in the codebase and explores the search space by mutating plans. Each step of the plan consists of a natural language intent and symbols from the codebase that can be used to implement the intent. The user request along with the detailed plan serves as an enriched query that provides necessary context from the codebase to any downstream coding system to convert the plan to code. Our plan search benefits from test-time compute scaling and produces repo-grounded plans without requiring code execution.  
    }
    \label{fig:teaser}
\end{figure}
\begin{figure}
    \centering
    \includegraphics[width=1.0\linewidth]{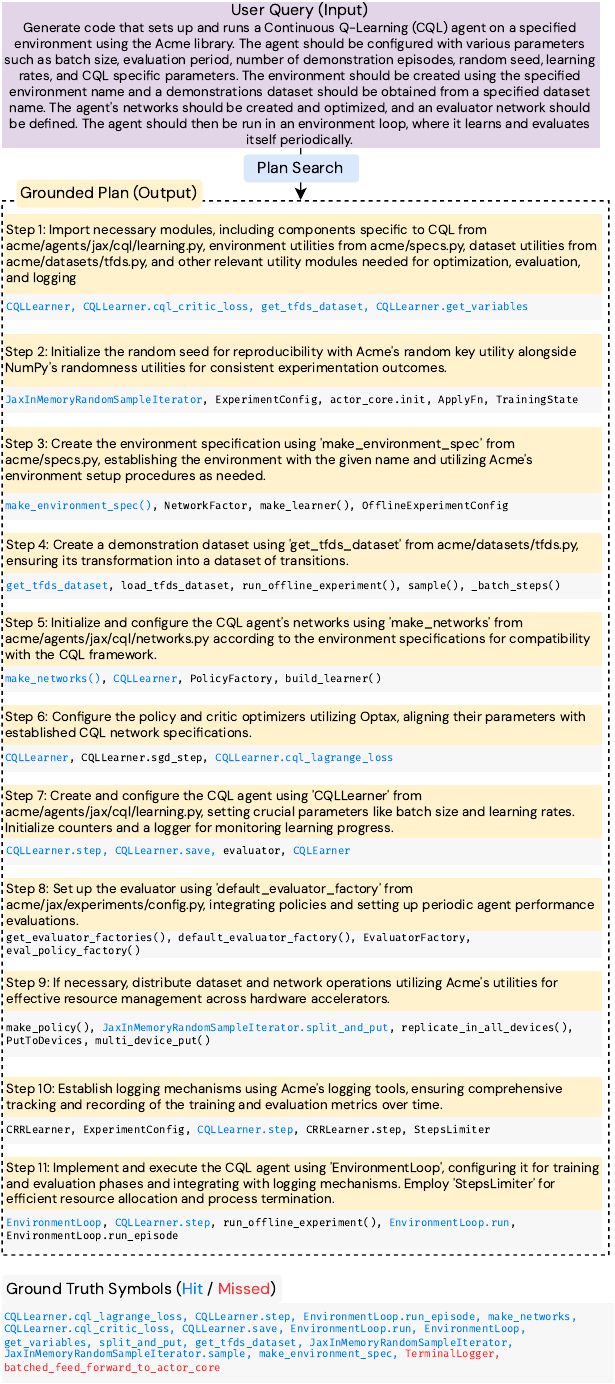}
    \caption{\small{\textbf{A repo-grounded plan created by \method~on a query from \lca}}. Each plan step consists of a natural language intent with top-5 symbols retrieved from the codebase that might be useful for implementing the step.}
    \label{fig:full-plan}
\end{figure}
Code generation systems powered by LLMs are routinely tasked with writing new code given an existing large codebase. One approach to conditioning an LLM's generation on a repository is to utilize its working memory by concatenating all files in the codebase into a massive prompt. This is an inefficient use of finite context, because many programming tasks require only a small fraction of all symbols (functions, classes, global variables etc) in the codebase. Recent investigation also shows that leading LLMs are effective at utilizing less than $20\%$ of their context lengths with a sharp decline in performance with increasing reasoning complexity \cite{babilong}. Can we do better?

Human programmers are able to understand complex codebases with a much smaller biological working memory. They achieve this by decomposing the target task into smaller steps and then using code search tools to identify relevant symbols (functions, classes etc) to use in each step. This is often an iterative process where the programmer interacts with the codebase to develop a realizable plan based on the symbols available in the codebase. The realizable plan can then be implemented.

In this work, we aim to replicate the ability of human programmers to iteratively search for a repo-grounded realizable plan to solve a user query given a codebase. This plan should wrap all relevant information from the codebase into a self-contained prompt which is human readable, editable and can be handed off to any code generation LLM or coding assistant for translation into code. 

Specifically, given a codebase and a user query that requires writing code using symbols in the codebase, we aim to find a repo-grounded plan with multiple steps. Each step consists of a natural language intent and the symbols from the codebase that may be used to implement or \textit{realize} the plan. A good plan is complete, concise, and faithful to the original query. These plans along with the original user query can be viewed as an enriched query that provides all relevant context from the codebase with detailed steps on how to solve the user query. 

Since the space of all plans is semi-structured and massive, \method~formulates repo-grounded plan search as an LLM-guided tree-search problem. Each node in the tree represents a plan. Each step of the search process involves identifying the most promising node to expand and create children or successors of the node by mutating the plan. The mutation aims to make the successors more accurate, repo-grounded and realizable.  When the search budget is exhausted the best plan is returned to the user. \textit{Importantly, \method~does not require executing any code. }

The design space of \method~consists of four key components - successor  function to mutate plans, a symbol retriever to ground intents to symbols in the codebase, a tree-traversal algorithm that decides the order in which to expand the nodes, and a plan ranker to select the most promising node to expand or to identify the best plan from the available nodes. We use the challenging \lca~benchmark~\cite{longcodearena} to thoroughly explore the design space of repo-grounded plan-search.

Our contributions include: (i) demonstrating the utility of repo-grounded plans for code-use (~\cref{tab:system-level-comparison-tokens}, ~\cref{fig:weak-llm},~\cref{fig:hard-tasks}); (ii) formulating execution-free repo-grounded planning as LLM-guided tree search using an intent to symbol grounding function (Sec.~\ref{sec:method}); (iii) elucidating and studying the design space of repo-grounded plan search (~\cref{fig:mutation-ablation}, ~\cref{fig:search-comparison},~\cref{tab:scoring-ablation}); and (iv) demonstrating that plan search allows code-use to benefit from gains by scaling test-time compute (~\cref{fig:mutation-ablation} and~\cref{fig:search-comparison}).

\section{Related Work}
\begin{figure}
    \centering
    \includegraphics[width=1.0\linewidth]{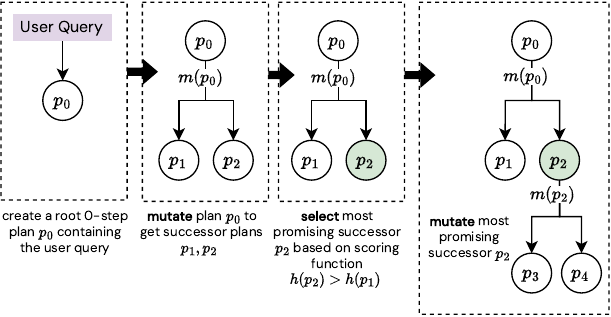}
    \vspace{-4mm}
    \caption{\textbf{Overview of plan search}. Each node in the tree is a repo-grounded plan. At every time step, a node is chosen for growing the tree and successors are created by mutating the chosen plan. We use an LLM to implement the successor function.}
    \label{fig:traversal-and-ranking}
\end{figure}
\noindent\textbf{Repository-grounded code generation.} Existing work on repo-level code generation has explored two distinct directions. One line of work focuses on building software engineering agents that can edit real-world codebases to solve Github issues. The challenge here involves understanding multiple files and coordinating edits across those files while executing unit tests to validate these changes. A popular benchmark for this paradigm is SWE-bench \cite{swe_bench} with several systems inching towards the performance of human engineers \cite{swe_agent, agentless}. 

Code-use is an alternate paradigm that involves using a codebase as a library to write new code to solve a user's query. This requires the code generation system to discover the relevant symbols (functions, classes, variables etc.) in the codebase, understand the syntax and function of these symbols, and write code using these symbols to solve the task. CodeNav~\cite{codenav} is a code-use agent that iteratively interacts with a keyword-based retrieval environment and an execution environment to solve the user's query. CodeNav repurposed tool-use benchmarks \cite{m3tooleval,mnms,api_bank} to evaluate code-use by providing the agent with the codebase implementing the tools instead of tool prompts. However, given the limited number of tools, and the simplicity of tools (simple functions) and user queries, these repurposed benchmarks fail to test the LLMs on the challenges of real-world code-use. 

In this work, we focus on the code-use scenario while using the recently released \lca~(LCA) \cite{longcodearena} benchmark. Specifically, we use the library-based code generation challenge in the LCA benchmark suite which curates tasks using example scripts found in prominent Github repositories. Since these examples scripts are provided by the library authors to demonstrate using their library for real tasks, LCA tasks present a significantly more challenging and realistic test-bed for code-use than previous code-use evaluations. 

\noindent\textbf{Plan search for code generation.} Recent work has shown the benefits of plan search for competitive programming tasks that require general knowledge of a programming language and its primitives. PlanSearch \cite{planning_improves_code_generation} demonstrates that searching over natural language plans before generating code leads to more diverse solutions and better performance. While both PlanSearch and \method~requires searching for plans that decompose a user query into a sequence of simpler steps, we further need to constrain search to the space of realizable plans i.e. plans where all steps can be implemented using the target codebase. 

\noindent\textbf{Test-time search for code generation.} Several systems such as AlphaCode\cite{alphacode}, CodeTree\cite{codetree}, and CodeMonkeys\cite{code_monkeys,large_language_monkeys} have demonstrated impressive performance on code generation tasks by scaling test-time compute to search in the space of programs while using execution feedback to guide the search. Similar to code search, AlphaGeometry\cite{alphageometry} utilizes neural-guided tree spearch to explore the solution space of geometric theorems represented using a formal geometric language with rich verifiers for validating each step. While our work also uses search for code generation, we search in the space of repo-grounded plans without execution feedback or formal verifiers. Nonetheless, we show that plans produced by our approach provide necessary context from the target codebase for the task of repo-grounded code generation.

\section{Method}\label{sec:method}
\begin{figure}
    \centering
    \includegraphics[width=1.0\linewidth]{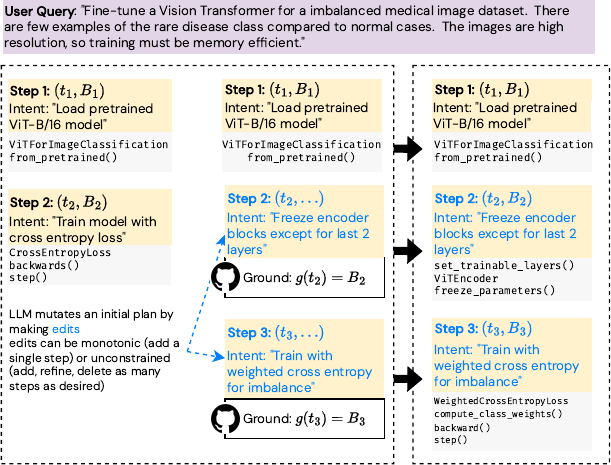}
    \vspace{-5mm}
    \caption{\small{\textbf{Mutation and grounding.}} The successor function $\mutationFn$ mutates a plan (left-most column) to generate new plans (right-most column). For each modified intent ($\intent_2$ and $\intent_3$), the grounding function $\intentToSymbols$ maps the intent to symbols that might be used to implement the intent ($\relevantSymbols_2$ and $\relevantSymbols_3$).}
    \label{fig:mutation-and-grounding}
\end{figure}

\paragraph{Overview} We formulate repository-grounded planning as a search problem over the space $\AllPlans$ of possible plans. Given the large space of possible plans and the semi-structured nature of plans, we employ a tree-based search algorithm using an LLM to guide the search process (\cref{fig:traversal-and-ranking}). Nodes in the tree represent the set of candidate plans explored so far with children created from parents via mutation using a successor function $\mutationFn: \AllPlans \rightarrow \mathbb{P}(\AllPlans)$ where $\mathbb{P}(\AllPlans)$ denotes the power set of $\mathcal{P}$. The search process begins at the root node which consists of the high-level user query as the initial plan $\plan_0 \in \AllPlans$. At each step, a node is selected for mutation using the scoring function $\rankingFn: \AllPlans \rightarrow \mathbb{R}$ (or using node expansion order as in depth-first search) and the successors are added to the set of candidates to be considered for future expansions. The process is repeated up to a user specified budget (nodes expanded), and the best plan (identified using $\rankingFn$) is returned. 

While this search procedure is applicable to a wide range of planning problems, we focus on choosing the space of plans, and appropriate mutation and scoring functions for the application of planning for repo-conditioned code generation. We conceptualize a plan $\plan \in \AllPlans$ as a sequence of steps $\planSequence$ where each step $\planStep_i$ is a tuple $(\intent, \relevantSymbols)$ consisting of a natural language intent $\intent \in \AllStrings$ and a set $\relevantSymbols \subseteq \AllSymbols$ of symbols relevant to implementing the intent chosen from the set $\AllSymbols$ of all symbols found in the codebase. To obtain relevant symbols, we construct a grounding function $\intentToSymbols: \AllStrings \rightarrow \mathbb{P}(\AllSymbols)$ that maps natural language intents to relevant symbols. Having relevant symbols from the target codebase as part of the plan is what makes these plans repo-grounded and realizable. 

For the successor function $\mutationFn: \AllPlans \rightarrow \mathbb{P}(\AllPlans)$, we explore two variants: an incremental version that only adds a new step to an existing plan, and an unconstrained version that can modify any part of the plan. Both variants use a language model to propose modifications to intents while using the grounding function to map intents to relevant symbols in the codebase.

Finally, we consider two scoring functions for ranking nodes: a heuristic function that encourages symbol diversity, and an LLM-based scoring function. For uninformed search like depth-first search, the scoring function is used solely for selecting the best plan after search completion, while for informed search like best-first search, the scoring function is also used to select the best node for expansion.

\begin{table}[t]
\caption{Notation used throughout this paper}
\label{tab:notation}
\resizebox{\columnwidth}{!}{
\begin{tabular}{ll}
\toprule
Notation & {\small Meaning} \\
\midrule
$\AllSymbols$ & {\small Set of all symbols (functions, classes, methods) in codebase} \\
$\AllStrings$ & {\small Set of all finite character strings (natural language)} \\
$\intent \in \AllStrings$ & {\small Natural language intent for a plan step} \\
$\relevantSymbols \subseteq \AllSymbols$ & {\small Set of relevant symbols for implementing a step} \\
$\planStep = \stepTuple$ & {\small Plan step: a tuple of intent and relevant symbols} \\
$\plan = \planSequence$ & {\small Plan: sequence of plan steps} \\
$\AllPlans$ & {\small Space of all possible plans} \\
$\intentToSymbols: \AllStrings \rightarrow \mathbb{P}(\AllSymbols)$ & {\small Maps intents to relevant symbols where $\mathbb{P}$ is the power set} \\
$\mutationFn: \AllPlans \rightarrow \mathbb{P}(\mathcal{P})$ & {\small Generates set of possible next plans or successors} \\
$\rankingFn: \AllPlans \rightarrow \mathbb{R}$ & {\small Scoring function for ranking plans} \\
\bottomrule
\end{tabular}
}
\vspace{-5mm}
\end{table}

\subsection{Successor Function and Grounding}
\label{sec:successors_and_grounding}
The successor function $\mutationFn: \AllPlans \rightarrow \mathbb{P}(\mathcal{P})$ determines how we explore the space of possible plans. For a given plan $\plan = \planSequence$ where each $\planStep_i = (\intent_i, \relevantSymbols_i)$, the successor function must generate new plans while attempting to ensure each step remains grounded in symbols $\AllSymbols$ available in the codebase.

\subsubsection{Successor Function Variants}

We consider two choices of the successor function:

\noindent\textbf{Monotonic.} The monotonic successor function $\mutationFn_m$ preserves all steps in the parent plan, only mutating the plan by adding new steps. Formally, given a plan $\plan = [(\intent_1, \relevantSymbols_1), ..., (\intent_n, \relevantSymbols_n)]$, $\mutationFn_m(\plan)$ generates plans of the form $[(\intent_1, \relevantSymbols_1), ..., (\intent_n, \relevantSymbols_n), (\intent_{n+1}, \relevantSymbols_{n+1})]$ where $\intent_{n+1}$ is a new intent and $\relevantSymbols_{n+1} \subseteq \AllSymbols$ contains the symbols needed to implement it. This ensures the search progressively builds longer plans while maintaining previously discovered steps, evocative of monotonic relaxations in planning \citep{Bonet2001,McDermott1999,Hoffmann2001}. 

\noindent\textbf{Unconstrained.} An unconstrained successor function $\mutationFn_u$ may perform arbitrary modifications to any part of the plan. For a plan $\plan$, $\mutationFn_u(\plan)$ can generate plans with modified, deleted, or reordered steps, while maintaining the requirement that each step $(\intent, \relevantSymbols)$ is grounded ($\relevantSymbols \subseteq \AllSymbols$). This allows the search to escape local optima by making dramatic changes to plans, similar to mutation operators in evolutionary search for planning \cite{evolution_vs_tree_search,10.1145/2463372.2463413}.

Both successor functions are implemented using an LLM with appropriate prompts (\cref{sec:successor-function-prompts}). The number of successors or branching factor is a crucial hyper parameter that allows us to control the allocation of the test-time compute budget -- a larger branching factor allows a greater exploration of the plan space $\AllPlans$. Given a branching factor of $f$, we sample $f$ times from the LLM (GPT-4o) to generate $f$ successors.

\subsubsection{Plan Grounding}
\label{sec:plan-grounding}
To guide the successor function and aid node scoring (for ranking), we need to ground each step intent in symbols found in the codebase that might be used to implement each step. This is achieved through the grounding function $\intentToSymbols: \AllStrings \rightarrow \mathbb{P}(\AllSymbols)$ which maps a natural language intent $\intent$ to relevant symbols in the codebase $B\subseteq \mathcal{B}$. This is challenging due to the semantic gap between high-level natural language intents and low-level code implementations \cite{ModalityGap}. Rather than attempting direct intent-to-code matching, we bridge this gap through an intermediate representation.

We use a retrieval-based approach for implementing the $\intentToSymbols$ that reduces the challenging intent-to-code grounding problem to an easier intent-to-intent matching problem. For each symbol $b \in \AllSymbols$, we use a lightweight language model to generate synthetic intents that describe potential uses of the symbol (e.g., "symbol $b$ can be used to..."). These synthetic intents are then embedded into a vector space using an embedding model $e: \AllStrings \rightarrow \mathbb{R}^d$. Given a plan step intent $\intent$, the grounding function retrieves the synthetic intents with the highest cosine similarity and returns the corresponding symbols. We return top-5 symbols after de-duplicating matches to the same symbol via alternate synthetic intents. We use GPT-4o-mini to generate the synthetic intents and text-embedding-3-large to compute intent embeddings. (Examples in \cref{tab:synthetic-intents-example}).

Having grounded symbols as part of the plan allows the successor function to identify infeasible or unrealizable steps in the plan and modify them. Similarly, the scoring function uses grounded symbols to score realizable plans more favorably than plans with unrealizable steps.

\subsection{Scoring Function and Exploration Strategies}
The scoring function $\rankingFn: \AllPlans \rightarrow \mathbb{R}$ plays two crucial roles in our approach. First, it enables informed search algorithms (e.g. best-first search) by guiding exploration toward promising regions of the plan space. Second, it allows selecting the most promising plans to pass to downstream code generation, even when using uninformed search strategies like depth-first search which simply rely on node expansion order via a stack data structure to determine the next plan to mutate.

Designing an effective ranking function is challenging because we need to impose an ordering over the plan space that correlates with two key properties: (1) the likelihood that a plan achieves the user's intent, and (2) the feasibility of implementing each step with the grounded symbols. Unlike traditional planning scenarios where the scoring function emerges naturally from the environment, we must construct a scoring function that can evaluate plans without executing them.

\subsubsection{Scoring Function Variants}\label{sec:scoring_functions}
\noindent\textbf{Symbol Diversity Scorer} implements a heuristic based on symbol coverage. For a plan $\plan = [(\intent_1, \relevantSymbols_1), ..., (\intent_n, \relevantSymbols_n)]$:

\begin{equation}
\rankingFn_{\text{sym}}(\plan) = |\bigcup_{i=1}^n \relevantSymbols_i|
\end{equation}

This rewards plans that incorporate a diverse set of symbols from the codebase, based on the intuition that effective plans likely require integrating multiple components. While simple, this approach provides a baseline that encourages thorough exploration of available functionality.

\noindent\textbf{Decomposed Likert Scorer} draws inspiration from recent work showing that decomposing evaluation into fine-grained criteria improves assessment reliability \cite{lmunit}. We construct a scoring function that evaluates both plan-level and step-level properties using a large language model as a judge. Given the user query, the plan, and symbol definitions, we ask an LLM to produce the following judgement scores one a 7-point Likert scale \cite{likert}: 
\begin{itemize}
\item A plan-level accuracy score $l_\plan$ assessing whether the plan solves the user request
\item Step-level feasibility scores ${l_1, ..., l_n}$ evaluating whether each step intent $t_i$ is realizable with the grounded symbols $B_i$
\end{itemize}

During informed search, we aggregate these scores into a single value to pick the next node for exploration:
\begin{equation}
\rankingFn_{\text{likert}}(\plan) = \frac{1}{2}\cdot\left(l_\plan + \frac{1}{n}\sum_{i=1}^n l_i\right)
\end{equation}

However, for final plan selection after search completes, we empirically found that a hierarchical sorting approach is more effective. Plans are first sorted by their plan-level score $l_\plan$, with ties broken by the average step-level score $\frac{1}{n}\sum_{i=1}^n l_i$. This two-level sorting ensures we prioritize plans that are likely to achieve the user's intent while using step-level feasibility as a secondary criterion. 

\textbf{Oracle Scorer.} For some of our ablations, we use symbol recall ($\%$ of ground truth symbols in generated plans) to score plans. Since this requires reference code this is not a practical setting but allows us to study the effect of components like successor function and tree-search algorithms on plan search performance in a controlled setting. 

\subsubsection{Exploration Strategies}
Our framework allows for plugging in any tree-search algorithm to guide the exploration of plan space $\AllPlans$. We primarily use best-first search in our experiments to make use of the scoring function for informed exploration while using depth-first search as an uninformed search baseline. We do not use breadth-first search because it is particularly wasteful for monotonic successor function as it spends most of its search budget on early stage incomplete plans. We leave more complex algorithms like MCTS for future work. 
\begin{table}
\caption{\small{\textbf{Comparing our plan based code generation to alternative approaches.}} Approaches are sorted by average amount of context usage. Using a fraction of the context, Plan Search is competitive with adding the entire codebase into the LLM context and significantly outperforms ReAct based planning.   
}
\label{tab:system-level-comparison-tokens}
\resizebox{\linewidth}{!}{
\begin{tabular}{@{}lrll@{}}
\toprule
 & & \multicolumn{2}{c}{Overlap Score} \\
\cmidrule(l{0.5em}r{0.5em}){3-4}
Context Fill & Avg. Tokens & Best-of-5 & Average \\ \midrule
Instruction Only & 250 & 42.3 & 32.2 \\
ReAct & 4,831 & 47.2 (\textcolor{blue}{+5.0}) & 39.8 (\textcolor{blue}{+7.6}) \\
Plan Search (ours) & 5,473 & 53.9 (\textcolor{blue}{+11.7}) & 48.0 (\textcolor{blue}{+15.8}) \\
\midrule
\rowcolor{gray!20}\textcolor{gray}{Full Repo} & 121,262 & 58.7 (\textcolor{gray}{+16.5}) & 49.9 (\textcolor{gray}{+17.6}) \\
\bottomrule
\end{tabular}
}
\vspace{-2mm}
\end{table}
\section{Experiments}
\textbf{Benchmark.} We evaluate our plans and code generated from our plans using the \lca~(LCA) benchmark\footnote{LCA contains multiple tracks. We use ``Library-based Code Generation'' track.}. Unlike traditional code generation benchmarks that focus on self-contained competition style programming problems that only require knowledge of a programming language and its primitives, \lca~tasks require understanding and using an external codebase to solve the user query. Each task provides a user query, the codebase required to implement the solution, and a reference solution making it more suitable for library-style code-use evaluations compared to the popular SWE-Bench which focuses on editing the codebase itself instead of using it as a library.  

\textbf{Metrics.} To evaluate the quality of generated code, we use the API Overlap metric introduced in \lca. This metric measures the recall of library symbols in the reference solution within the generated code. Specifically, for both the reference and generated code, we extract all symbols from their abstract syntax trees using the \texttt{tree-sitter} library and filter for symbols that belong to the target repository. The overlap score is then computed as the percentage of reference symbols that appear in the generated code. This metric captures how well the generated code utilizes the appropriate functionality from the codebase, while being robust to superficial differences in implementation.

For all experiments involving code generation, we sample 5 solutions per configuration and report both the best-of-5 and average overlap scores. The best-of-5 score reflects the best performance achieved by the downstream LLM that translates our plans to code in 5 independent tries, while the average score indicates average across those tries or expected performance. When evaluating plans directly (without code generation), we measure plan recall -- the percentage of symbols from the reference solution that appear in the retrieved symbols of any plan step.

\textbf{Evaluation Overview.} We systematically evaluate our system components: (\cref{sec:system_level_comparison}) compares the effectiveness of generating code conditioned on our grounded plans with alternatives, (\cref{sec:successor-function-ablation}) the impact of different successor functions, (\cref{sec:traversal-ablation}) the choice of search strategy, and (\cref{tab:scoring-ablation}) the effectiveness of different ranking functions. We conclude by demonstrating that our searched plans can help weaker language models match the performance of stronger models on these tasks in \cref{sec:enhancing-other-llms} and enable progress on hard tasks where even a frontier model (GPT-4o) with a context window full of repository context makes little progress (\cref{sec:hard-longcodearena-tasks}). Codegen prompts are in \cref{sec:codegen-prompts}.
\subsection{System-Level Comparisons}\label{sec:system_level_comparison}
\label{sec:system-level-comparison}

\begin{figure}
    \centering
    \includegraphics[width=\linewidth]{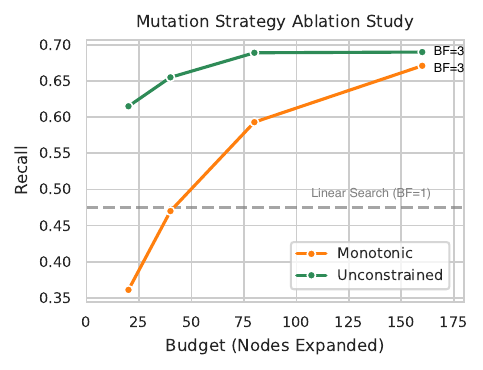}
    \vspace{-5mm}
    \caption{\textbf{\textcolor{ForestGreen}{Unconstrained} mutation outperforms \textcolor{Orange}{monotonic} mutation, especially at lower budgets}. Here, we show the symbol recall (\% of ground truth symbols in the generated plans) of each mutation strategy using best-first search with the oracle scoring function and branching factor of $3$. This figure also illustrates gains from scaling test-time compute (by increasing budget).}
    \label{fig:mutation-ablation}
\end{figure}
First, we evaluate plan search as part of the end-to-end system that generates code given a user query. This compares the overlap scores for the code generated from our plans to alternative approaches. We use GPT-4o (128K context window) for both plan search and for generating code from plans. Specifically, we compare the following approaches:

\begin{itemize}
    \item \textbf{Instruction Only:} The model receives only the user query with no additional context from the codebase.
    \item \textbf{ReAct:} In our plan search framework, a ReAct baseline is equivalent to setting branching factor to $1$ with a monotonic successor function resulting in a linear chain instead of a tree of plans. The final plan is provided as context for code generation.
    \item \textbf{Plan Search:} Our approach using unconstrained successor function, informed best-first search (branching factor=3, budget=80), and the symbol diversity scorer. The resulting plan is provided as context for code generation. For both ReAct and PlanSearch we use a maximum tree depth (chain length for ReAct) of $20$.
    \item \textbf{Full Repo:} The entire repository is provided as context for code generation to establish an upper bound that fully utilizes the LLM's context window.
\end{itemize}

As shown in Table~\ref{tab:system-level-comparison-tokens}, the instruction-only baseline achieves an overlap score of 42.3\% (best) and 32.2\% (average), demonstrating that models have some ability to guess appropriate API usage from instructions alone. ReAct (linear search) improves upon this baseline (+5.0\% best, +7.6\% average) by actively searching the codebase.
Tree-structured plan search outperforms both baselines (+11.7\% best, +15.8\% average over instruction-only) while using only 4.3\% of the context window. Notably, this performance approaches that of using the full repository as context, despite using less than 5\% of the available context budget. This demonstrates that our search can construct highly effective plans that capture precisely parts of the codebase needed to solve each task.

\subsection{Successor Function Ablation}
\label{sec:successor-function-ablation}
We evaluate how the choice of successor function impacts plan search performance. We compare two variants of the successor function: (i) \textit{monotonic}, which can only add new steps and (ii) \textit{unconstrained}, which can modify any part of the plan (see Section~\ref{sec:successors_and_grounding}). To isolate the effect of the successor function, we fix other components of the system: informed search (best-first) with branching factor of 3, maximum tree depth of 20, and use an oracle ranker that scores plans based on their recall of ground truth symbols for the user query. We use GPT-4o to guide the plan search and vary the search budget (nodes expanded) from 20 to 160.

Figure~\ref{fig:mutation-ablation} shows plan's symbol recall (percentage of ground truth symbols found) as a function of search budget. First, note that for both successor functions, performance improves with increasing search budget. This re-affirms the role of scaling test-time compute to improving reasoning performance. Next, we see that that unconstrained mutation consistently outperforms monotonic mutation across all budgets, with the gap being particularly pronounced at lower budgets (+30\% at budget=20). The unconstrained successor achieves its peak performance with fewer steps as compared to monotonic successor suggesting more efficient exploration of the plan space through non-monotonic changes. We also show the performance of ReAct (linear search with a monotonic successor as described in Section~\ref{sec:system_level_comparison}) for reference. Tree search significantly outperforms linear search, regardless of the choice of successor function.

\subsection{Traversal Ablation}
\label{sec:traversal-ablation}
\begin{figure}
    \centering
    \includegraphics[width=\linewidth]{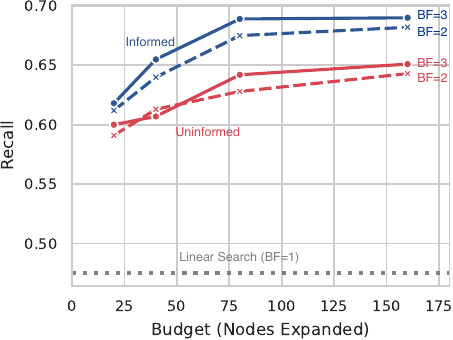}
    \vspace{-3mm}
    \caption{\small{\textbf{Comparison of Search Strategies.}} {\color[HTML]{2E5894}Informed} (best-first) search outperforms {\color[HTML]{D64550}uninformed} (depth-first) and linear search strategies and performance improves with branching factor (BF), especially for informed search.}
    \label{fig:search-comparison}
\end{figure}

We now investigate how different search strategies and branching factors affect plan quality. We compare informed search (best-first) against uninformed search (depth-first) as well as linear search. For this experiment, we use the unconstrained successor function and budget of 160 with a maximum tree depth of 20. As in the previous experiment, we use an oracle ranker for informed search to establish an upper bound on achievable performance.

As shown in \cref{fig:search-comparison}, informed search significantly outperforms uninformed and linear search strategies with branching factor of 3 showing healthy gains over 2. This is because informed search makes a more efficient use of the compute budget by exploring more promising parts of the plan space $\AllPlans$. Both tree-search strategies do much better than linear search which does not explore alternative solutions.

\subsection{Scoring Function Ablation}
\label{sec:scoring-function-ablation}
\begin{table}
\centering
\caption{\small{\textbf{Comparing scoring functions}}. While diversity scorer scores slightly better on the overlap metrics, an LLM judge prefers the plans chosen by the Likert scorer suggesting higher fidelity of Likert-chosen plans to the original user query.}
\label{tab:scoring-ablation}
\begin{tabular}{llrr}
\toprule
 &  & \multicolumn{2}{c}{Overlap Score} \\
\cmidrule{3-4}
Scoring Fn. & Win Rate & Best-of-N & Average \\
\midrule
Diversity & 35\% & 49.8 & 41.9 \\
Likert & 65\% & 47.2 & 39.8 \\
\bottomrule
\end{tabular}
\end{table}
\begin{figure}
    \centering
    \includegraphics[width=\linewidth]{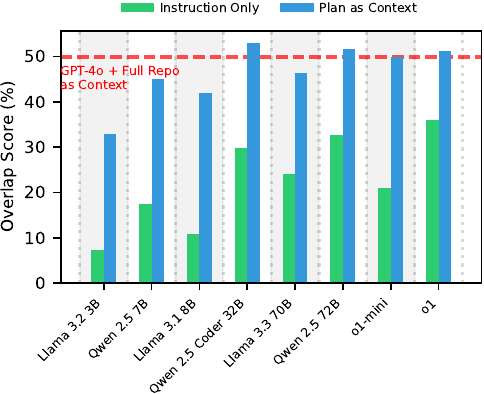}
    \vspace{-3mm}
    \caption{\small{\textbf{Our plans consistently improve performance across all models.}} Qwen 2.5 Coder 32B with our plans exceeds GPT-4o's full-repo performance despite conditioning on 120k fewer context tokens. Even models stronger than GPT-4o (e.g., O1) benefit from our GPT-4o-generated plans. The red line shows GPT-4o performance when given the full repository as context. 
    }
    \vspace{-2mm}
    \label{fig:weak-llm}
\end{figure}
Previous ablations used an oracle scorer to establish the potential of different search strategies. In practice we need to score plans without access to ground truth symbols. We compare our symbol diversity scorer and decomposed likert scorer from Section~\ref{sec:scoring_functions}. For this evaluation, we first generate candidate plans using uninformed depth-first search with a budget of 160 and the monotonic successor function and then use each scoring function to select the best plan among these candidates. The selected plan for each scoring function is then given to GPT-4o to generate the code.

We evaluate the scoring functions in two ways. First, we do a pairwise comparison between code generated by using the two scoring functions using an LLM judge. Given the code generated from both scoring functions, we ask an LLM judge (GPT-4o) to pick the code that better matches the reference code. We then compute the win rate of each scoring function. To ensure reliable evaluation, we collect 6 judgments per pair and take a majority vote. Second, we compute overlap scores using reference code. Table~\ref{tab:scoring-ablation} shows that while the diversity scorer achieves slightly higher overlap scores (49.8\% best-of-N vs 47.2\%), the plans selected using the Likert scorer are preferred by LLM judge in pairwise comparisons (65\% win rate). We hypothesize that this might be due to the Likert scorer's ability to pick plans with more accurate decomposition of the user query into step level intents than the diversity scorer which does not consider the fidelity of the plans to the original user query. 
\subsection{Enhancing Other LLMs with Searched Plans}
\label{sec:enhancing-other-llms}

\begin{figure}
    \centering
    \includegraphics[width=\linewidth]{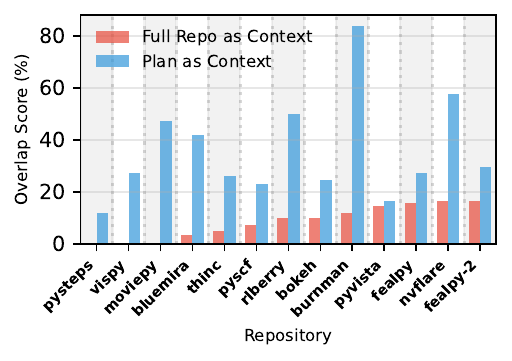}
    \vspace{-8mm}
    \caption{\small{\textbf{Our plans enable progress on hard tasks where even full-repo context performed poorly.}} Conditioning on tree-searched plans shows gains on the hardest 10\% of tasks where GPT-4o with full-repo context performed poorly.}
    \label{fig:hard-tasks}
    \vspace{-3mm}
\end{figure}
Next, we examine whether plans searched by our system can enhance the performance of other language models, particularly smaller open-source models. We test a range of open models models from 3B to 70B parameters, including both general-purpose LLMs (Llama, Qwen) and models specifically fine-tuned for code (Qwen Coder), as well as OpenAI's reasoning models (O1, O1-mini). For each model, we compare performance with instruction only (i.e. no repo context) and performance with instructions augmented by the plans generated by our approach as repo context.

Figure~\ref{fig:weak-llm} shows several striking results. First, our searched plans consistently and significantly improve performance across all models tested. The magnitude of improvement is particularly dramatic for smaller models - Llama 3.2 3B improves from 7.4\% to 32.9\% overlap score when given our plans, while Qwen 2.5 7B improves from 17.5\% to 45.0\%. Qwen 2.5 Coder 32B with our plans (53.0\% overlap) outperforms GPT-4o with full repository access (49.9\% overlap), despite conditioning on 95\% less context. This suggests that our plans are providing a more efficient form of context than raw repository content, enabling smaller models to match or exceed the performance of much larger ones.

Our plans improve performance even for models that are stronger than the one used to generate the plans. For instance, O1, which achieves a 36.1\% score with instructions alone (significantly better than GPT-4o's instruction-only performance), sees substantial gains from our GPT-4o-generated plans, improving to 51.1\%, while O1-mini improves from 21.0\% to 50.0\% overlap given our plans.

\subsection{Impact of Plans on Hard \lca~Tasks}
\label{sec:hard-longcodearena-tasks}

We analyze performance on the most challenging tasks in \lca~(the 10\% of tasks for which GPT-4o with the entire repository as context makes the least progress). Figure~\ref{fig:hard-tasks} compares providing the full repository as context (red) against using our tree-searched plans as context (blue). For each approach, we sample 5 programs conditioned on the context per task and show the average performance using the overlap metric.
On these challenging tasks, GPT-4o with full repository context (128K tokens) struggles significantly, achieving average overlap scores below 20\% across all repositories. In contrast, our tree-searched plans (unconstrained successor, branching factor=3) enable substantially better performance across all tasks. For instance, on the \texttt{burnman} task, our approach achieves an average overlap of 83.75\%, while full repository context manages only 12\%.

\section{Conclusion}

\method~automatically enriches user queries with repo-grounded plans found through execution-free tree search. Our system decomposes high-level requests into detailed plans where each step pairs natural language intent with relevant codebase symbols. Through experiments on LongCodeArena, we demonstrated that our plans: (1) are effective as context for code generation; (2) enable weaker models to match stronger models' performance; and (3) enable progress on challenging tasks where even frontier models with full repository context struggle. Our results show that grounded plan search is a promising direction for improving code-use while maintaining efficiency and interpretability.

\section*{Impact Statement}
This paper presents work whose goal is to advance the field
of Machine Learning. There are many potential societal
consequences of our work, none which we feel must be
specifically highlighted here.

\bibliography{references}
\bibliographystyle{icml2024}

\newpage
\appendix
\onecolumn
This appendix provides additional details, analyses and experimental results to supplement the main paper. \cref{sec:synthetic-intents-examples} presents qualiative examples of the synthetic intents used to ground plans in a codebase. \cref{sec:extended-hard-task-analysis} presents an in-depth analysis of our system's performance on challenging tasks from LongCodeArena. \cref{sec:successor-function-prompts} shows the complete prompts used for our successor functions. \cref{sec:scoring-prompts} details the prompts used for our scoring functions. \cref{sec:codegen-prompts} provides the prompts used for code generation across different experimental settings.
\section{Qualitative Examples of Synthetic Intents}
\label{sec:synthetic-intents-examples}

\begin{longtable}{p{3cm}p{6cm}p{6cm}}
\caption{\textbf{Qualitative Examples of Synthetic Intents}: We show randomly selected examples of the top 3 closest synthetic intents by embedding distance (using OpenAI's \texttt{text-embedding-3-large}) to a query intent. Each synthetic intent is generated conditioned on a symbol from the codebase (using \texttt{GPT-4o-mini}). To ground plan steps during plan search, the intent from the plan step is matched to synthetic intents and therefore to the symbols corresponding to the synthetic intents.} \\
\label{tab:synthetic-intents-example} \\
\toprule
Repository & Intent & Top-3 Closest Synthetic Intents (\texttt{Symbol}) \\
\midrule
\endfirsthead

\multicolumn{3}{l}{\emph{Continued from previous page}} \\
\toprule
Repository & Intent & Top-3 Closest Synthetic Intents (\texttt{Symbol}) \\
\midrule
\endhead

\midrule
\multicolumn{3}{r}{\emph{Continued on next page}} \\
\endfoot

\bottomrule
\endlastfoot

\multirow{3}{*}{pybamm} & Add a no SEI submodel. & \textit{I need to create a new discretisation instance without providing a mesh.} (\texttt{Discretisation}) \newline \textit{I want to create an isothermal thermal submodel for my simulation.} (\texttt{Isothermal}) \newline \textit{I want to assign parameter values to a specific model.} (\texttt{process\_model}) \\
\cmidrule(l){2-3}
& Add a constant porosity submodel. & \textit{I want to create an instance of the constant concentration diffusion model.} (\texttt{ConstantConcentration}) \newline \textit{I want to create an isothermal thermal submodel for my simulation.} (\texttt{Isothermal}) \newline \textit{I want to initialize a constant concentration model for the diffusion process with specific parameters.} (\texttt{ConstantConcentration}) \\
\cmidrule(l){2-3}
& Add an isothermal thermal submodel. & \textit{I want to create an isothermal thermal submodel for my simulation.} (\texttt{Isothermal}) \newline \textit{I need to set up the Isothermal model for my thermal simulations.} (\texttt{Isothermal}) \newline \textit{I need to gather all temperature variables associated with the isothermal submodel.} (\texttt{get\_fundamental\_variables}) \\
\midrule
\multirow{3}{*}{dd4hep} & Set up a particle gun with specified parameters. & \textit{I want to set up a particle gun in the simulation to start generating particles.} (\texttt{setupGun}) \newline \textit{I need to configure the particle gun with specific parameters such as name, particle type, and energy level.} (\texttt{setupGun}) \newline \textit{I want to customize the position and multiplicity settings for the particle gun in the simulation.} (\texttt{setupGun}) \\
\cmidrule(l){2-3}
& Set up a tracker for the simulation. & \textit{I want to set up a tracking field for my particle simulation using a specific configuration.} (\texttt{setupTrackingFieldMT}) \newline \textit{I want to set up the construction of the detector in the simulation.} (\texttt{detectorConstruction}) \newline \textit{I want to configure the tracking field setup for my Geant4 simulation.} (\texttt{setupTrackingField}) \\
\cmidrule(l){2-3}
& Set up event actions for particle printing. & \textit{I am looking to set up a generator action for particle generation in my application.} (\texttt{GeneratorAction}) \newline \textit{I want to set up a particle gun in the simulation to start generating particles.} (\texttt{setupGun}) \newline \textit{I would like to use the tracking action functionality to monitor particle tracks in my experiment.} (\texttt{TrackingAction}) \\
\midrule
\multirow{3}{*}{fealpy} & Create a uniform time mesh for the simulation. & \textit{I want to generate the initial mesh for my 2D time harmonic solver.} (\texttt{init\_mesh}) \newline \textit{I want to ensure that the mesh is refined uniformly to improve simulation accuracy.} (\texttt{init\_mesh}) \newline \textit{I want to create a uniform triangular mesh to use in my analysis.} (\texttt{init\_mesh}) \\
\cmidrule(l){2-3}
& Solve the linear system to update the solution at the current time step. & \textit{I need to update my solution by solving the linear system after applying Dirichlet boundary conditions.} (\texttt{solve}) \newline \textit{I need to iterate through time steps and update my model's solutions.} (\texttt{time\_integration}) \newline \textit{I need to update the state of my model variables after solving the system.} (\texttt{solve}) \\
\cmidrule(l){2-3}
& Advance to the next time level in the time mesh. & \textit{I want to progress the time in my algorithm by moving to the next time level.} (\texttt{next\_time\_level}) \newline \textit{I want to advance to the next time level in the simulation.} (\texttt{next\_time\_level}) \newline \textit{I want to progress to the subsequent time level in the timeline.} (\texttt{next\_time\_level}) \\
\midrule
\multirow{3}{*}{nplab} & Create an experiment class that involves a shutter and a spectrometer. & \textit{I want to initialize a new shutter instance in my experiment setup.} (\texttt{Shutter}) \newline \textit{I want to prepare a shutter for my nanophotonics experiments.} (\texttt{Shutter}) \newline \textit{I need to construct a shutter object to manage exposure times.} (\texttt{Shutter}) \\
\cmidrule(l){2-3}
& Initialize and display the GUI application. & \textit{I want to initialize a new GUI widget that will display a plot.} (\texttt{Widget}) \newline \textit{I want to initialize the user interface for the spectrometers in the application.} (\texttt{\_init\_ui}) \newline \textit{I need to initialize a GUI component that displays spectrometer controls.} (\texttt{SpectrometersUI}) \\
\cmidrule(l){2-3}
& Define properties for irradiation time and wait time. & \textit{I need to configure the integration time and delay settings for my spectrometer to ensure accurate time series measurements.} (\texttt{update\_time\_series\_params}) \newline \textit{I need to expose the instrument for a set amount of time and ensure it blocks until the exposure completes.} (\texttt{expose}) \newline \textit{I want to specify the duration for which the spectrometer should collect data during a measurement.} (\texttt{set\_integration\_time}) \\
\midrule
\multirow{3}{*}{python-sc2} & Manage drones to gather minerals if vespene gas is above a certain threshold or Zergling speed upgrade is pending. & \textit{I want to check if my unit is currently gathering resources from a mineral field or vespene geyser.} (\texttt{is\_gathering}) \newline \textit{I need to identify the units that are engaged in gathering minerals or vespene.} (\texttt{gathering}) \newline \textit{I need to direct a unit to gather either minerals or gas for my economy.} (\texttt{gather}) \\
\cmidrule(l){2-3}
& Research Zergling speed upgrade if conditions are met. & \textit{I need to queue an upgrade research for my unit.} (\texttt{research}) \newline \textit{I need to determine how fast my unit can move considering the effects of any active upgrades.} (\texttt{calculate\_speed}) \newline \textit{I want to start researching an upgrade if the necessary tech building is ready.} (\texttt{research}) \\
\cmidrule(l){2-3}
& Draw a creep pixelmap for debugging purposes. & \textit{I want to check if there is creep on a specific grid point in the game.} (\texttt{has\_creep}) \newline \textit{I want to output debug information by drawing a box around a game unit.} (\texttt{debug\_box2\_out}) \newline \textit{I want to draw a visual line between two points in my game for debugging purposes.} (\texttt{debug\_line\_out}) \\
\midrule
\multirow{3}{*}{basilisk} & Initialize and execute the simulation within the scenario execution function. & \textit{I need to initialize the simulation and prepare all modules for execution.} (\texttt{SimBaseClass}) \newline \textit{I want to execute a simulation by assigning the appropriate execution function.} (\texttt{setExecutionFunction}) \newline \textit{I need to prepare my simulation for execution by initializing all required data structures and parameters.} (\texttt{SimBaseClass}) \\
\cmidrule(l){2-3}
& Import necessary modules and set up file paths for the simulation. & \textit{I need to initialize the simulation and prepare all modules for execution.} (\texttt{SimBaseClass}) \newline \textit{I want to configure my simulation environment with the correct paths and logger setup on initialization.} (\texttt{SimBaseClass}) \newline \textit{I need to ensure that all modules in the simulation are properly self-initialized.} (\texttt{InitializeSimulation}) \\
\cmidrule(l){2-3}
& Define a function to execute the simulation scenario, including configuring stop time and initializing the simulation. & \textit{I want to define an execution function that will run my simulation instance.} (\texttt{setExecutionFunction}) \newline \textit{I want to define the parameters for running a simulation, including the creation and execution functions.} (\texttt{SimulationParameters}) \newline \textit{I want to define how long my simulation should run by setting the stop time.} (\texttt{ConfigureStopTime}) \\

\end{longtable}

\section{More Analysis of Hard Tasks on LongCodeArena}
\label{sec:extended-hard-task-analysis}
\begin{figure}
    \centering
    \includegraphics{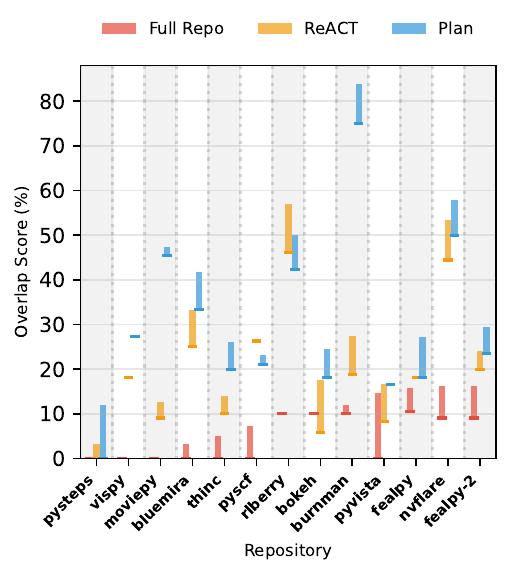}
    \caption{Performance comparison on the most challenging LongCodeArena tasks (bottom 10th percentile by full-repo performance). For each repository, we show the performance of GPT-4o when given: the full repository as context (red), ReACT-generated plans (orange), or our tree-searched plans (blue). Bars show average performance across 5 samples, while the bottom of error bars indicates worst-case performance. Our tree-searched plans (using unconstrained successor function and branching factor=3) consistently outperform both baselines, with worst-case performance often exceeding the baselines' average performance. All scores are API overlap percentages measuring alignment with reference solutions.}
    \label{fig:hard-tasks-spread}
\end{figure}
To better understand the robustness of our approach, we analyze worst-case and average-case performance on the most challenging tasks in LongCodeArena (bottom 10th percentile by full-repository performance). Figure~\ref{fig:hard-tasks-spread} compares three approaches using GPT-4o as the code generator: providing the full repository as context (red), using ReACT-generated plans (orange), and using our tree-searched plans (blue). For each approach, we sample 5 solutions per task and show both the average performance (top of bars) and minimum performance (bottom of error bars) using the API overlap metric.

Our approach consistently outperforms ReACT-style planning, showing better average performance on 11 out of 13 tasks. More importantly, the worst-case performance with our plans (indicated by the bottom of the blue bars) often exceeds the average performance of both baselines, suggesting that tree-searched plans lead to more reliable code generation. This is particularly evident in repositories like moviepy, where our approach's minimum performance (45.45\%) far exceeds both the ReACT average (12.73\%) and full repository average (0\%).

These results demonstrate that systematic tree search produces more robust plans than either naive context inclusion or linear planning approaches, particularly on challenging tasks where standard approaches struggle to make progress.
\section{Successor Function Prompts}
\label{sec:successor-function-prompts}
\begin{figure}
    \centering
    \includegraphics[width=0.8\textwidth]{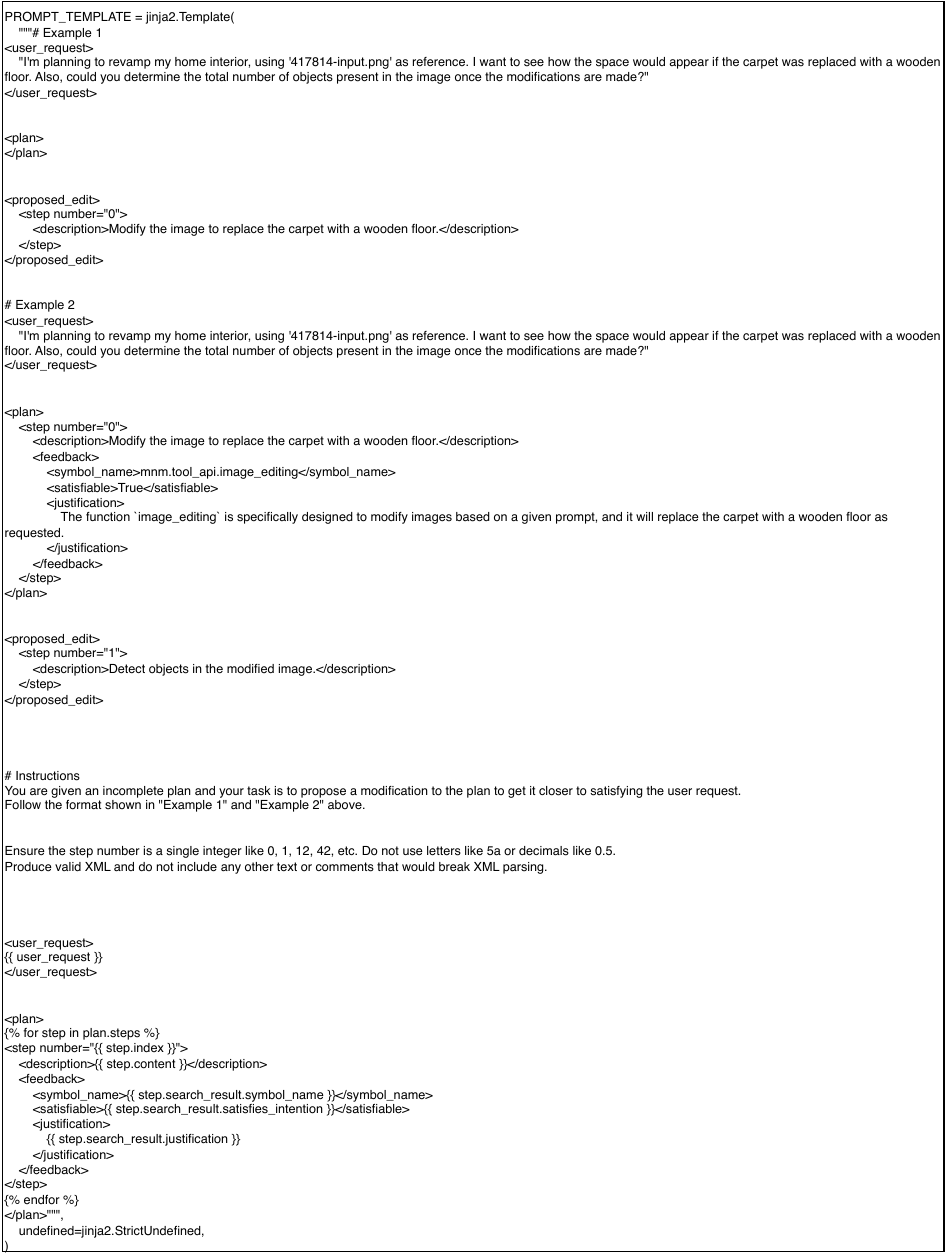}
    \caption{\textbf{Monotonic Successor Function Prompt}}
    \label{fig:monotonic-successor-prompt}
\end{figure}
\begin{figure}
    \centering
    \includegraphics[width=0.8\textwidth]{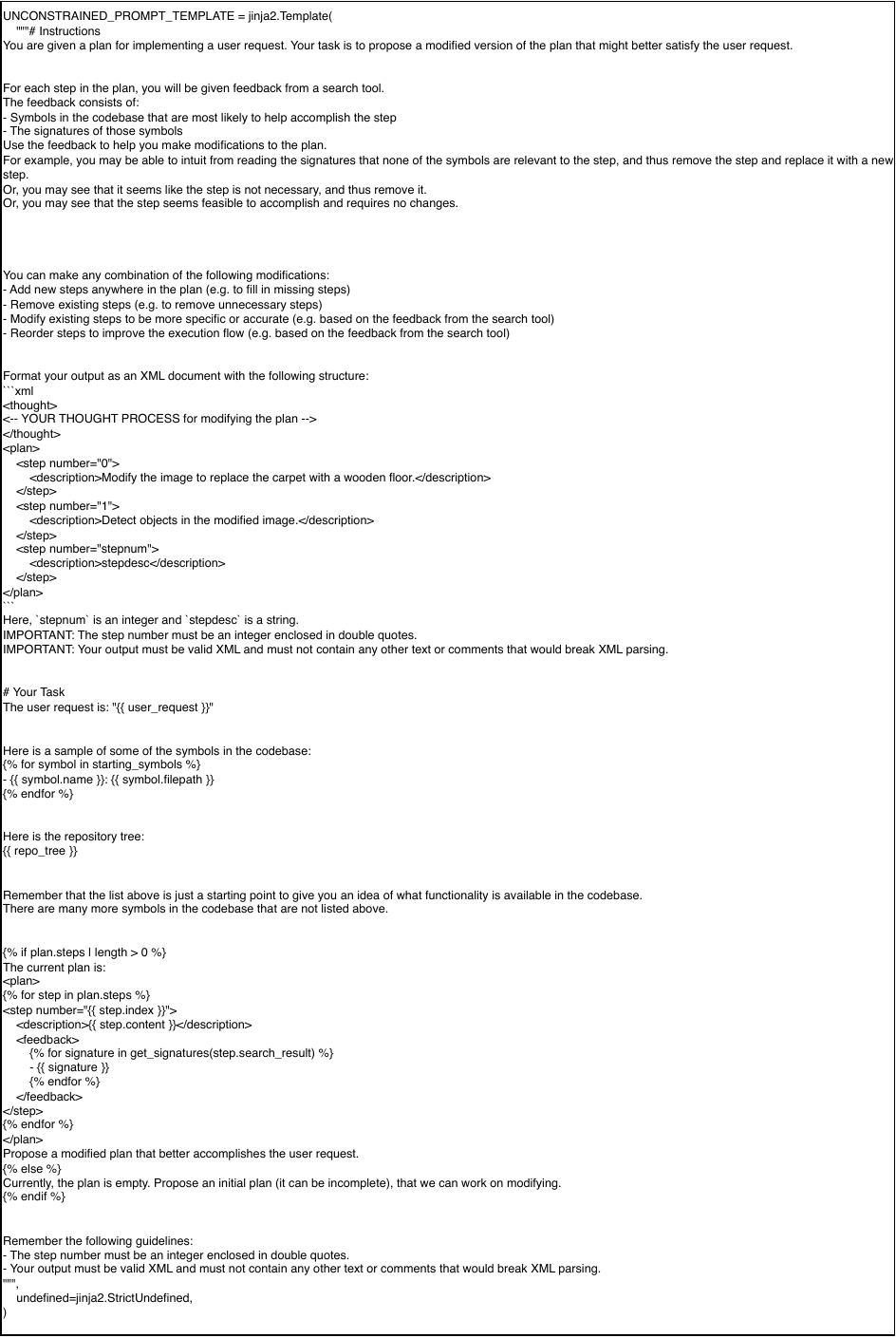}
    \caption{\textbf{Unconstrained Successor Function Prompt}}
    \label{fig:unconstrained-successor-prompt}
\end{figure}
Our successor functions rely on prompts to guide the LLM in mutating plans. \cref{fig:monotonic-successor-prompt} shows the prompt template used for the monotonic successor function, which can only add new steps while preserving existing ones. \cref{fig:unconstrained-successor-prompt} shows the prompt for the unconstrained successor function, which can modify any part of the plan. 
\section{Scoring Function Prompts}
\label{sec:scoring-prompts}
\begin{figure}
    \centering
    \includegraphics[width=0.8\linewidth]{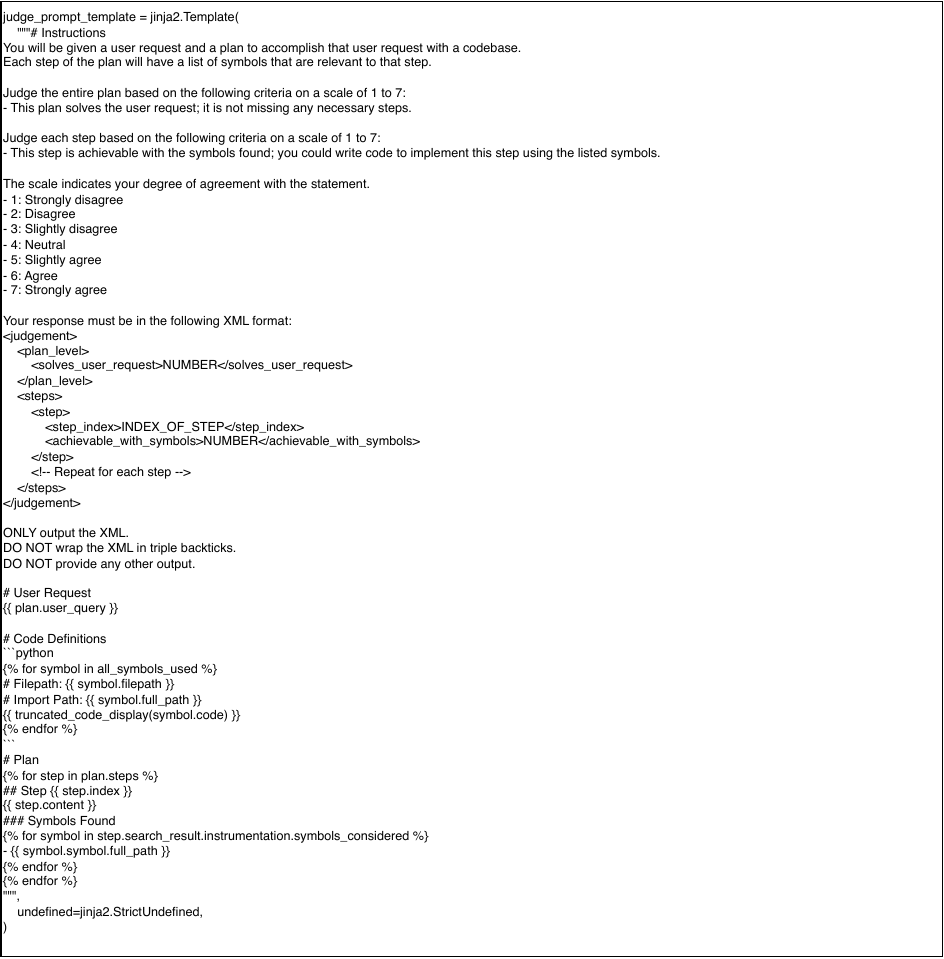}
    \caption{\textbf{Likert Scoring Function Prompt}}
    \label{fig:likert-judge-prompt}
\end{figure}
The Likert scoring function uses the prompt shown in \cref{fig:likert-judge-prompt} to evaluate plans. The prompt breaks down evaluation into two aspects: (1) whether the overall plan achieves the user's intent and (2) whether each step is feasible with its retrieved symbols. The scoring is done on a 7-point Likert scale.
\section{Code Generation Prompts}
\label{sec:codegen-prompts}
\begin{figure}
    \centering
    \includegraphics[width=0.8\linewidth]{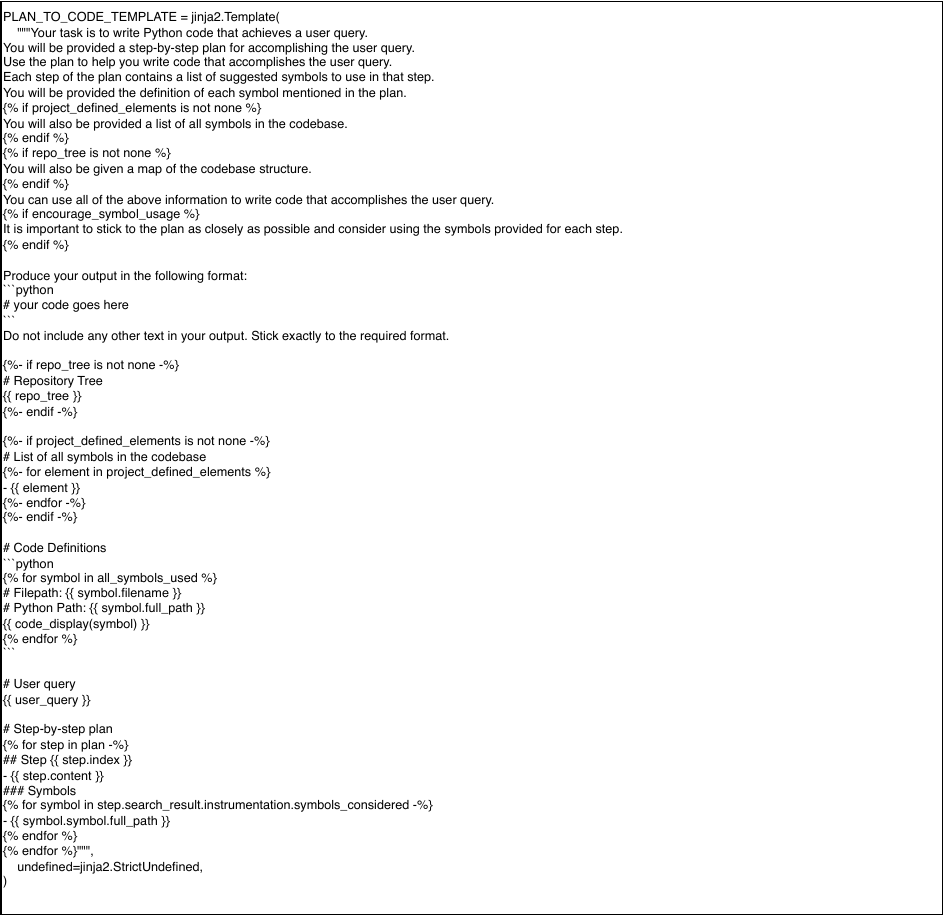}
    \caption{\textbf{Plan-based Code Generation Prompt}}
    \label{fig:plan-based-codegen-prompt}
\end{figure}
\begin{figure}
    \centering
    \includegraphics{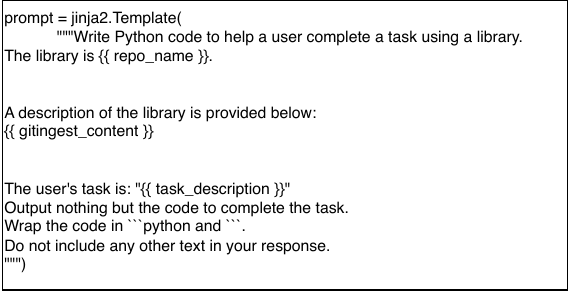}
    \caption{\textbf{Full-repository Context Code Generation Prompt}}
    \label{fig:full-repo-codegen-prompt}
\end{figure}
\begin{figure}
    \centering
    \includegraphics{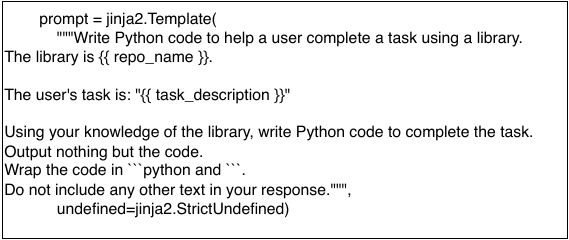}
    \caption{\textbf{Instruction-only Code Generation Prompt}}
    \label{fig:instruction-only-codegen-prompt}
\end{figure}
The code generation prompts are designed to evaluate different approaches to providing repository context. \cref{fig:plan-based-codegen-prompt} shows how we present plans as structured context for code generation. \cref{fig:full-repo-codegen-prompt} demonstrates the full-repository baseline approach where the entire codebase is provided as context. \cref{fig:instruction-only-codegen-prompt} shows the minimal instruction-only setting which provides no additional context beyond the user query.


\end{document}